\documentclass[11pt]{article}
\usepackage{acl-hlt2011}
\usepackage{times}
\usepackage{latexsym}
\usepackage{amsmath}
\usepackage{multirow}
\usepackage{url}

\title{Translating from Morphologically Complex Languages:\\
       A Paraphrase-Based Approach}

\author{Preslav Nakov\\
  Department of Computer Science\\
  National University of Singapore\\
  13 Computing Drive\\
  Singapore 117417\\
  {\tt nakov@comp.nus.edu.sg}
  \And
  Hwee Tou Ng\\
  Department of Computer Science\\
  National University of Singapore\\
  13 Computing Drive\\
  Singapore 117417\\
  {\tt nght@comp.nus.edu.sg}}

\date{}

\begin{document}
\maketitle
\begin{abstract}
We propose a novel approach to translating from a morphologically complex language.
Unlike previous research,
which has targeted word inflections and concatenations,
we focus on the pairwise relationship between morphologically related words,
which we treat as \emph{potential paraphrases}
and handle using paraphrasing techniques at the word, phrase, and sentence level.
An important advantage of this framework
is that it can cope with derivational morphology,
which has so far remained largely beyond the capabilities
of statistical machine translation systems.
Our experiments translating from Malay,
whose morphology is mostly derivational,
into English
show significant improvements over rivaling approaches
based on five automatic evaluation measures
(for 320,000 sentence pairs; 9.5 million English word tokens).

\end{abstract}

%%%%%%%%%%%%%%%%%%%%%%%%%%%%%%%%%%%%%%%%%%%%%%%%%%
\section{Introduction}
\label{sect:intro}

Traditionally, statistical machine translation (SMT) models
have assumed that the \emph{word} should be the basic token-unit of translation,
thus ignoring any word-internal morphological structure.
This assumption can be traced back to the first word-based models of IBM \cite{brown93mathematic},
which were initially proposed for two languages with limited morphology: French and English.
While several significantly improved models have been developed since then,
including
phrase-based \cite{koehn-smt},
hierarchical \cite{Chiang:hiero},
treelet \cite{quirk-menezes-cherry:2005:ACL},
and syntactic \cite{Galley:al:syntaxMT} models,
they all preserved the assumption that words should be atomic.

Ignoring morphology was fine as long as the main research interest remained focused
on languages with limited (e.g., English, French, Spanish) or minimal (e.g., Chinese) morphology.
Since the attention shifted to languages like Arabic, however,
the importance of morphology became obvious
and several approaches to handle it have been proposed.
Depending on the particular language of interest,
researchers have paid attention to \emph{word inflections} and \emph{clitics},
e.g., for Arabic, Finnish, and Turkish,
or to \emph{noun compounds},
e.g., for German.
However, \emph{derivational morphology}
has not been specifically targeted so far.

In this paper, we propose
a paraphrase-based approach to translating from a morphologically complex language.
Unlike previous research,
we focus on the pairwise relationship between morphologically related wordforms,
which we treat as \emph{potential paraphrases},
and which we handle using paraphrasing techniques at various levels: word, phrase, and sentence level.
An important advantage of this framework
is that it can cope with various kinds of morphological wordforms,
including derivational ones.
We demonstrate its potential on Malay,
whose morphology is mostly derivational.

The remainder of the paper is organized as follows:
Section \ref{sec:malay:morpho} gives an overview of Malay morphology,
Section \ref{sec:method} introduces our paraphrase-based approach to translating from morphologically complex languages,
Section \ref{sec:eval} describes our dataset and our experimental setup,
Section \ref{sec:discuss} presents and analyses the results,
and Section \ref{sect:relwork} compares our work to previous research.
Finally, Section \ref{sec:conclusion} concludes the paper
and suggests directions for future work.

\section{Malay Morphology and SMT}
\label{sec:malay:morpho}

Malay is an Astronesian language, spoken by about 180 million people.
It is official in Malaysia, Indonesia, Singapore, and Brunei,
and has two major dialects,
sometimes regarded as separate languages,
which are mutually intelligible,
but occasionally differ in orthography/pronunciation and vocabulary:
Bahasa Malaysia (\emph{lit.}~`language of Malaysia')
and Bahasa Indonesia (\emph{lit.} `language of Indonesia').

Malay is an agglutinative language with very rich morphology.
Unlike other agglutinative languages such as Finnish, Hungarian, and Turkish,
which are rich in both inflectional and derivational forms,
Malay morphology is mostly derivational.
Inflectionally,\footnote{\emph{Inflection} is variation in the form of a word
that is obligatory in some given grammatical context.
For example, \emph{plays}, \emph{playing}, \emph{played}
are all inflected forms of the verb \emph{play}.
It does not yield a new word and cannot change the part of speech.}
Malay is very similar to Chinese:
there is no grammatical gender, number, or tense,
verbs are not marked for person, etc.

In Malay, new words can be formed by the following three morphological processes:

\begin{itemize}
    \item \textbf{Affixation}, i.e., attaching affixes, which are not words themselves, to a word.
These can be
prefixes (e.g., \emph{ajar}/`teach' $\rightarrow$ \emph{\textbf{pel}ajar}/`student'),
suffixes (e.g., \emph{ajar} $\rightarrow$ \emph{ajar\textbf{an}}/`teachings'),
circumfixes (e.g., \emph{ajar} $\rightarrow$ \emph{\textbf{peng}ajar\textbf{an}}/`lesson'), and
infixes (e.g., \emph{gigi}/`teeth' $\rightarrow$ \emph{ge\textbf{ri}gi}/`toothed blade').
Infixes only apply to a small number of words and are not productive.

    \item \textbf{Compounding}, i.e., forming a new word by putting two or more existing words together.
    For example, \emph{kereta}/`car' + \emph{api}/`fire' make \emph{kereta api} and \emph{keretapi}
    in Bahasa Indonesia and Bahasa Malaysia, respectively, both meaning `train'.
    As in English, Malay compounds are written separately,
    but some stable ones like \emph{kerjasama}/`collaboration'
    (from \emph{kerja}/`work' and \emph{sama}/`same') are concatenated.
    Concatenation is also required when a circumfix is applied to a compound,
    e.g., \emph{ambil alih}/`take over' (\emph{ambil}/`take' + \emph{alih}/`move')
    is concatenated to form \emph{\textbf{peng}ambilalih\textbf{an}}/`takeover'
    when targeted by the circumfix \emph{peng-$\ldots$-an}.

    \item \textbf{Reduplication}, i.e., word repetition.
    In Malay, reduplication requires using a dash.
It can be
full (e.g., \emph{pelajar-pelajar}/`students'),
partial (e.g., \emph{adik-beradik}/`siblings', from \emph{adik}/`younger brother/sister'), and
rhythmic (e.g., \emph{gunung-ganang}/`mountains', from the word \emph{gunung}/`mountain').
\end{itemize}

Malay has very little inflectional morphology,
It also has some \textbf{clitics}\footnote{A \emph{clitic}
is a morpheme that has the syntactic characteristics of a word,
but is phonologically bound to another word.
For example, \emph{'s} is a clitic in \emph{The Queen of England's crown}.},
which are not very frequent and are typically spelled concatenated to the preceding word.
For example, the politeness marker \emph{lah}
can be added to the command \emph{duduk}/`sit down'
to yield \emph{duduk\textbf{lah}}/`please, sit down',
and the pronoun \emph{nya} can attach to \emph{kereta}
to form \emph{kereta\textbf{nya}}/`his car'.
Note that clitics are not affixes,
and clitic attachment is not a word derivation or a word inflection process.

Taken together, affixation, compounding, reduplication, and clitic attachment
yield a rich variety of wordforms,
which cause data sparseness issues.
Moreover, the predominantly derivational nature of Malay morphology
limits the applicability of standard techniques such as
(1)~removing some/all of the source-language inflections,
(2)~segmenting affixes from the root, and
(3)~clustering words with the same target translation.
For example, if \emph{\textbf{pel}ajar}/`student' is an unknown word
and lemmatization/stemming reduces it to \emph{ajar}/`teach',
would this enable a good translation?
Similarly, would segmenting\footnote{The prefix \emph{peN}
suffers a nasal replacement of the archiphoneme \emph{N} to become \emph{pel} in \emph{pelajar}.}
\emph{\textbf{pel}ajar} as \emph{peN+ ajar},
i.e., as `person doing the action' + `teach',
make it possible to generate `student' (e.g., as opposed to `teacher')?
Finally, if affixes tend to change semantics so much,
how likely are we to find morphologically related wordforms that share the same translation?
Still, there are many good reasons to believe that morphological processing should help SMT for Malay.

Consider \emph{affixation},
which can yield words with similar semantics
that can use each other's translation options,
e.g.,
\emph{\textbf{di}ajar}/`be taught (intransitive)'
and \emph{\textbf{di}ajarkan}/`be taught (transitive)'.
However, this cannot be predicted from the affix, e.g.,
compare \emph{minum}/`drink (verb)' -- \emph{minum\textbf{an}}/`drink (noun)'
and
\emph{makan}/`eat' -- \emph{makan\textbf{an}}/`food'.

Looking at \emph{compounding},
it is often the case that the semantics of a compound is a specialization of the semantics of its head,
and thus the target language translations available for the head could be usable to translate the whole compound,
e.g., compare \emph{kerjasama}/`collaboration' and \emph{kerja}/`work'.
Alternatively, it might be useful to consider a segmented version of the compound,
e.g., \emph{kerja sama}.

\emph{Reduplication}, among other functions, expresses plural,
e.g., \emph{pelajar-pelajar}/`students'.
Note, however, that it is not used when a quantity or a number word is present, e.g.,
\emph{dua pelajar}/`two students' and \emph{banyak pelajar}/`many students'.
Thus, if we do not know how to translate \emph{pelajar-pelajar},
it would be reasonable to consider the translation options for \emph{pelajar}
since it could potentially contain among its translation options the plural `students'.

Finally, consider \emph{clitics}.
In some cases, a clitic could express a fine-grained distinction such as politeness,
which might not be expressible in the target language;
thus, it might be feasible to simply remove it.
In other cases, e.g., when it is a pronoun,
it might be better to segment it out as a separate word.

\section{Method}
\label{sec:method}

We propose a \emph{paraphrase-based approach} to Malay morphology,
where we use paraphrases
at three different levels:
word, phrase, and sentence level.

First, we transform each development/testing Malay sentence into a \emph{word lattice},
where we add simplified \emph{word-level paraphrasing} alternatives for each morphologically complex word.
In the lattice, each alternative $w'$ of an original word $w$ is assigned the weight of $\mathrm{Pr}(w'|w)$,
which is estimated using pivoting over the English side of the training bi-text.
Then, we generate \emph{sentence-level paraphrases} of the training Malay sentences,
in which exactly one morphologically complex word is substituted by a simpler alternative.
Finally, we extract additional Malay phrases from these sentences,
which we use to augment the phrase table with additional translation options
to match the alternative wordforms in the lattice.
We assign each such additional phrase $p'$ a probability $\max_{p}\mathrm{Pr}(p'|p)$,
where $p$ is a Malay phrase that is found in the original training Malay text.
The probability is calculated using \emph{phrase-level pivoting}
over the English side of the training bi-text.

%Below, we first present the morphological analysis we perform.
%We then describe our paraphrases and their weights
%for each level of translation.

\subsection{Morphological Analysis}

Given a Malay word, we build a list of morphologically simpler words
that could be derived from it;
we also generate alternative word segmentations:

\begin{itemize}
\item[(a)] words obtainable by affix stripping\\ e.g., \emph{pelajaran} $\rightarrow$ \emph{pelajar}, \emph{ajaran}, \emph{ajar}
\item[(b)] words that are part of a compound word\\ e.g., \emph{kerjasama} $\rightarrow$ \emph{kerja} %\emph{sama}
\item[(c)] words appearing on either side of a dash\\ e.g., \emph{adik-beradik} $\rightarrow$ \emph{adik}, \emph{beradik}
\item[(d)] words without clitics\\ e.g., \emph{keretanya} $\rightarrow$ \emph{kereta}
\item[(e)] clitic-segmented word sequences\\ e.g., \emph{keretanya} $\rightarrow$ \emph{kereta nya}
\item[(f)] dash-segmented wordforms\\ e.g., \emph{aceh-nias} $\rightarrow$ \emph{aceh - nias}
\item[(g)] combinations of the above.
\end{itemize}

The list is built by reversing the basic morphological processes in Malay:
(a)~addresses affixation,
(b)~handles compounding,
(c)~takes care of reduplication, and
(d) and (e)~deal with clitics.
Strictly speaking, (f) does not necessarily model a morphological process:
it proposes an alternative tokenization,
but this could make morphological sense too.
% e.g., for compounds with internal dashes.

Note that (g) could cause potential problems when interacting with (f),
e.g., \emph{adik-beradik} would become \emph{adik - beradik}
and then by (a) it would turn into \emph{adik - adik},
which could cause the SMT system to generate two separate translations for the two instances of \emph{adik}.
To prevent this, we forbid the application of (f) to reduplications.
Taking into account that reduplications can be partial,
we only allow (f)
if $\frac{|LCS(l,r)|}{\min(|l|,|r|)} < 0.5$,
where
$l$ and $r$ are the strings to the left and to the right of the dash, respectively,
$LCS(x,y)$ is the longest common character subsequence, not necessarily consecutive, of the strings $x$ and $y$,
and $|x|$ is the length of the string $x$.
For example, $LCS$(\emph{adik},\emph{beradik})=\emph{adik},
and thus, the ratio is 1 ($\geq 0.5$) for \emph{adik-beradik}.
Similarly, $LCS$(\emph{gunung},\emph{ganang})=\emph{gnng},
and thus, the ratio is 4/6=0.67 ($\geq 0.5$) for \emph{gunung-ganang}.
However, for \emph{aceh-nias}, it is 1/4=0.25,
and thus (f) is applicable.

As an illustration,
here are the wordforms we generate for \emph{adik-beradiknya}/`his siblings':
\emph{adik},
\emph{adik-beradiknya},
\emph{adik-beradik nya},
\emph{adik-beradik},
\emph{beradiknya},
\emph{beradik nya},
\emph{adik nya}, and
\emph{beradik}.
And for \emph{berpelajaran}/`is educated', we build the list:
\emph{berpelajaran},
\emph{pelajaran},
\emph{pelajar},
\emph{ajaran}, and
\emph{ajar}.
Note that the lists do include the original word.

To generate the above wordforms,
we used two morphological analyzers:
a freely available Malay lemmatizer \cite{Baldwin06opensource},
and an in-house re-implementation of the Indonesian stemmer described in \cite{Adriani:2007}.
Note that these tools' objective is to return a single lemma/stem,
e.g., they would return \emph{adik} for \emph{adik-beradiknya},
and \emph{ajar} for \emph{berpelajaran}.
However, it was straightforward to modify them
to also output the above intermediary wordforms,
which the tools were generating internally anyway when looking for the final lemma/stem.
Finally, since the two modified analyzers had different strengths and weaknesses,
we combined their outputs to increase recall.

\subsection{Word-Level Paraphrasing}
\label{sec:word-level}

We perform word-level paraphrasing of the Malay sides of the development and the testing bi-texts.
%adding morphologically-derived alternatives.

First, for each Malay word,
we generate the above-described list of morphologically simpler words and alternative word segmentations;
we think of the words in this list as \emph{word-level paraphrases}.
Then, for each development/testing Malay sentence,
we generate a lattice encoding all possible paraphrasing options for each individual word.

We further specify a weight for each arc.
We assign 1 to the original Malay word $w$,
and $\mathrm{Pr}(w'|w)$ to each paraphrase $w'$ of $w$,
where $\mathrm{Pr}(w'|w)$ is the probability that $w'$ is a \emph{good paraphrase} of $w$.
Note that multi-word paraphrases, e.g., resulting from clitic segmentation,
are encoded using a sequence of arcs;
in such cases, we assign $\mathrm{Pr}(w'|w)$ to the first arc, and 1 to each subsequent arc.

We calculate the probability $\mathrm{Pr}(w'|w)$
using the training Malay-English bi-text,
which we align at the word level using IBM model 4 \cite{brown93mathematic},
and we observe which English words $w$ and $w'$ are aligned to.
More precisely, we use \emph{pivoting}
to estimate the probability $\mathrm{Pr}(w'|w)$ as follows:

\begin{center}
    $\mathrm{Pr}(w'|w) = \sum_i \mathrm{Pr}(w'|w,e_i)\mathrm{Pr}(e_i|w)$
\end{center}

Then, following \cite{callison06paraphrase,pivot},
we make the simplifying assumption that $w'$ is conditionally independent of $w$ given $e_i$,
thus obtaining the following expression:

\begin{center}
    $\mathrm{Pr}(w'|w) = \sum_i \mathrm{Pr}(w'|e_i)\mathrm{Pr}(e_i|w)$
\end{center}

We estimate the probability $\mathrm{Pr}(e_i|w)$
directly from the word-aligned training bi-text as follows:

\begin{center}
    $\mathrm{Pr}(e_i|w) = \frac{\#(w,e_i)}{\sum_j\#(w,e_j)}$
\end{center}

\noindent where $\#(x,e)$ is the number of times the Malay word $x$ is aligned to the English word $e$.

Estimating $\mathrm{Pr}(w'|e_i)$ cannot be done directly
since $w'$ might not be present on the Malay side of the training bi-text,
e.g., because it is a multi-token sequence generated by clitic segmentation.
Thus, we think of $w'$ as a pseudoword that stands for the union of all Malay words in the training bi-text
that are reducible to $w'$ by our morphological analysis procedure.
So, we estimate $\mathrm{Pr}(w'|e_i)$ as follows:

\begin{center}
    $\mathrm{Pr}(w'|e_i) = \mathrm{Pr}(\{v:w' \in forms(v)\}|e_i)$
\end{center}

\noindent where $forms(x)$ is the set of the word-level paraphrases\footnote{Note
that our paraphrasing process is directed:
the paraphrases are morphologically simpler than the original word.}
for the Malay word $x$.

Since the training bi-text occurrences of the words that are reducible to $w'$
are distinct, we can rewrite the above as follows:

\begin{center}
    $\mathrm{Pr}(w'|e_i) = \sum_{v : w' \in forms(v)} \mathrm{Pr}(v|e_i)$
\end{center}

Finally, the probability $\mathrm{Pr}(v|e_i)$ can be estimated using maximum likelihood:

\begin{center}
    $\mathrm{Pr}(v|e_i) = \frac{\#(v,e_i)}{\sum_u\#(u,e_i)}$
\end{center}

\subsection{Sentence-Level Paraphrasing}

In order for the word-level paraphrases to work,
there should be phrases in the phrase table
that could potentially match them.
For some of the words, e.g., the lemmata,
there could already be such phrases,
but for other transformations,
e.g., clitic segmentation,
this is unlikely.
Thus, we need to augment the phrase table with additional translation options.

One approach would be to modify the phrase table directly,
e.g., by adding additional entries, where one or more Malay words are replaced by their paraphrases.
This would be problematic since the phrase translation probabilities
associated with these new entries would be hard to estimate.
For example, the clitics, and even many of the intermediate morphological forms,
would not exist as individual words in the training bi-text,
which means that there would be no word alignments or lexical probabilities available for them.
%Moreover, at least in principle,
%we would need to re-normalize all other probabilities,
%while avoiding double-counting.

Another option would be to generate separate word alignments
for the original training bi-text
and for a version of it where the source (Malay) side has been paraphrased.
Then, the two bi-texts and their word alignments would be concatenated
and used to build a phrase table \cite{Dyer:noisier:channel:2007,dyer-muresan-resnik:2008:ACLMain,Dyer:lattices:MT:2009}.
This would solve the problems with the word alignments
and the phrase pair probabilities estimations
in a principled manner,
but it would require choosing for each word only one of the paraphrases available to it,
while we would prefer to have a way to allow all options.
Moreover, the paraphrased and the original versions of the corpus would be given equal weights,
which might not be desirable.
Finally, since the two versions of the bi-text would be word-aligned separately,
there would be no interaction between them,
which might lead to missed opportunities for improved alignments
in both parts of the bi-text \cite{nakov-ng:2009:EMNLP}.

We avoid the above issues by adopting a sentence-level paraphrasing approach.
Following the general framework proposed in \cite{Nakov:08},
we first create multiple paraphrased versions of the source-side sentences of the training bi-text.
Then, each paraphrased source sentence is paired with its original translation.
This augmented bi-text is word-aligned and a phrase table $T'$ is built from it,
which is merged with a phrase table $T$ for the original bi-text.
The merged table contains all phrase entries from $T$,
and the entries for the phrase pairs from $T'$ that are not in $T$.
Following \newcite{nakov-ng:2009:EMNLP},
we add up to three additional indicator features (taking the values 0.5 and 1)
to each entry in the merged phrase table, showing whether the entry came from
(1)~$T$ only,
(2)~$T'$ only, or
(3)~both $T$ and $T'$.
We also try using the first one or two features only.
We set all feature weights using minimum error rate training \cite{och03minimum},
and we optimize their number (one, two, or three) on the development dataset.\footnote{
         In theory, we should re-normalize the probabilities;
         in practice, this is not strictly required by the log-linear SMT model.}

Each of our paraphrased sentences differs from its original sentence by a single word,
which prevents combinatorial explosions:
on average, we generate 14 paraphrased versions per input sentence.
It further ensures that the paraphrased parts of the sentences will not dominate the word alignments or the phrase pairs,
and that there would be sufficient interaction at word alignment time
between the original sentences and their paraphrased versions.

\subsection{Phrase-Level Paraphrasing}

While our sentence-level paraphrasing
informs the decoder about the origin of each phrase pair (original or paraphrased bi-text),
it provides no indication about how good the phrase pairs from the paraphrased bi-text are likely to be.

Following \newcite{callison06paraphrase},
we further augment the phrase table with one additional feature
whose value is 1 for the phrase pairs coming from the original bi-text,
and $\max_{p}\mathrm{Pr}(p'|p)$
for the phrase pairs extracted from the paraphrased bi-text.
Here $p$ is a Malay phrase from $T$,
and $p'$ is a Malay phrase from $T'$ that does not exist in $T$
but is obtainable from $p$ by substituting one or more words in $p$
with their derivationally related forms generated by morphological analysis.
The probability $\mathrm{Pr}(p'|p)$ is calculated using phrase-level pivoting through English
in the original phrase table $T$ as follows (unlike word-level pivoting, here $e_i$ is an English \emph{phrase}):

\begin{center}
    $\mathrm{Pr}(p'|p) = \sum_i \mathrm{Pr}(p'|e_i)\mathrm{Pr}(e_i|p)$
\end{center}

We estimate the probabilities $\mathrm{Pr}(e_i|p)$ and $\mathrm{Pr}(p'|e_i)$
as we did for word-level pivoting,
except that this time we use the list of the phrase pairs extracted
from the original training bi-text,
while before we used IBM model 4 word alignments.
When calculating $\mathrm{Pr}(p'|e_i)$,
we think of $p'$ as the set of all possible Malay phrases $q$ in $T$
that are reducible to $p'$ by morphological analysis of the words they contain.
This can be rewritten as follows:

\begin{center}
    $\mathrm{Pr}(p'|e_i) = \sum_{q : p' \in par(q)} \mathrm{Pr}(q|e_i)$
\end{center}

\noindent where $par(q)$ is the set of all possible phrase-level paraphrases for the Malay phrase $q$.

The probability $\mathrm{Pr}(q|e_i)$ is estimated using maximum likelihood from the list of phrase pairs.
There is no combinatorial explosion here, since
the phrases are short and contain very few paraphrasable words.

\section{Experiments}
\label{sec:eval}

\begin{table*}[tbh]
\begin{center}
\begin{scriptsize}
%\begin{tabular}{@{ }ll@{ }@{ }@{ }l@{ }@{ }@{ }l@{ }@{ }@{ }l@{ }@{ }@{ }l@{ }@{ }@{ }l@{ }@{ }@{ }l@{ }@{ }@{ }l@{ }@{ }@{ }l@{ }}
\begin{tabular}{llllllllll}
\multicolumn{1}{r}{Number of sentence pairs} & \textbf{1K} & \textbf{2K} & \textbf{5K} & \textbf{10K} & \textbf{20K} & \textbf{40K} & \textbf{80K} & \textbf{160K} & \textbf{320K} \\
\multicolumn{1}{r}{Number of English words} & 30K & 60K & 151K & 301K & 602K & 1.2M & 2.4M & 4.7M & 9.5M\\
\hline
\hline
                       baseline & 23.81 & 27.43 & 31.53 & 33.69 & 36.68 & 38.49 & 40.53 & 41.80 & 43.02\\
\hline
                 lemmatize all & 22.67 & 26.20 & 29.68 & 31.53 & 33.91 & 35.64 & 37.17 & 38.58 & 39.68 \\
                               & -1.14 & -1.23 & -1.85 & -2.16 & -2.77 & -2.85 & -3.36 & -3.22 & -3.34 \\
\hline
         `noisier' channel model \cite{Dyer:noisier:channel:2007} & 23.27 & \textbf{28.42} & \textbf{32.66} & 33.69 & 37.16 & 38.14 & 39.79 & 41.76 & 42.77 \\
                               & -0.54 & \textbf{+0.99} & \textbf{+1.13} & +0.00 & \textbf{+0.48} & -0.35 & -0.74 & -0.04 & -0.25 \\
\hline
\hline
         lattice + sent-par (orig+lemma) & \textbf{24.71} & \textbf{28.65} & \textbf{32.42} & \textbf{34.95} & \textbf{37.32} & 38.40 & 39.82 & 41.97 & 43.36\\
                               & \textbf{+0.90} & \textbf{+1.22} & \textbf{+0.89} & \textbf{+1.26} & \textbf{+0.64} & -0.09 & -0.71 & +0.17 & +0.34 \\

\hline
                       lattice + sent-par & \textbf{24.97} & \textbf{29.11} & \textbf{33.03} & \textbf{35.12} &
                                           \textbf{37.39} & 38.73 & \textbf{41.04} & 42.24 & \textbf{43.52}\\
                                & \textbf{+1.16} & \textbf{+1.68} & \textbf{+1.50} & \textbf{+1.43} & \textbf{+0.71} & +0.24 & \textbf{+0.51} & \textbf{+0.44} & \textbf{+0.50}\\
\hline
          lattice + sent-par + word-par & \textbf{25.14} & \textbf{29.17} & \textbf{33.00} & \textbf{35.09} &
                                           \textbf{37.39} & 38.76 & \emph{40.75} & \textbf{42.23} & \textbf{43.58}\\
                                & \textbf{+1.33} & \textbf{+1.74} & \textbf{+1.47} & \textbf{+1.40} & \textbf{+0.71} & +0.27 & \emph{+0.22} & \textbf{+0.43} & \textbf{+0.56}\\

\hline
lattice + sent-par + word-par + phrase-par & \textbf{25.27} & \textbf{29.19} & \textbf{33.35} & \textbf{35.23} &
                                           \textbf{37.46} & \textbf{39.00} & \textbf{40.95} & \textbf{42.30} & \textbf{43.73}\\
                                & \textbf{+1.46} & \textbf{+1.76} & \textbf{+1.82} & \textbf{+1.54} & \textbf{+0.78} & \textbf{+0.51} & \textbf{+0.42} & \textbf{+0.50} & \textbf{+0.71} \\

\hline
\hline
\end{tabular}
\end{scriptsize}
\end{center}
\caption{
\label{table:eval:berita} \textbf{Evaluation results.}
    Shown are BLEU scores and improvements over the baseline (in \%)
    for different numbers of training sentences.
    Statistically significant improvements are in \textbf{bold} for $p < 0.01$ and in \emph{italic} for $p < 0.05$.}
\end{table*}

\subsection{Data}
\label{sec:data}

We created our Malay-English training and development datasets from data
that we downloaded from the Web and then sentence-aligned using various heuristics.
Thus, we ended up with 350,003 \emph{training} sentence pairs,
including
10.4M English and 9.7M Malay word tokens.
We further downloaded 49.8M word tokens of monolingual English text,
which we used for \emph{language modeling}.

For \emph{testing}, we used 1,420 sentences with 28.8K Malay word tokens,
which were translated by three human translators,
yielding translations of 32.8K, 32.4K, and 32.9K English word tokens, respectively.
For \emph{development}, we used 2,000 sentence pairs of 63.4K English and
58.5K Malay word tokens.

\subsection{General Experimental Setup}

First, we tokenized and lowercased all datasets: training, development, and testing.
We then built directed word-level alignments for the training bi-text
for English$\rightarrow$Malay and for Malay$\rightarrow$English
using IBM model 4 \cite{brown93mathematic},
which we symmetrized using the intersect+grow heuristic \cite{och03:asc}.
Next, we extracted phrase-level translation pairs of maximum length seven,
which we scored and used to build a phrase table
where each phrase pair is associated with the following five standard feature functions:
forward and reverse phrase translation probabilities,
forward and reverse lexicalized phrase translation probabilities,
and phrase penalty.

We trained a log-linear model using the following standard SMT feature functions:
trigram language model probability, word penalty,
distance-based distortion cost, and the five feature functions from the phrase table.
We set all weights on the development dataset
by optimizing BLEU \cite{BLEU}
% directly
using minimum error rate training \cite{och03minimum},
and we plugged them in a beam search decoder \cite{moses2007acl}
to translate the Malay test sentences to English.
Finally, we detokenized the output,
and we evaluated it against the three
%lowercased
reference translations.
% using BLEU.

\subsection{Systems}

Using the above general experimental setup,
we implemented the following baseline systems:

\begin{itemize}

\item \textbf{baseline}.
        This is the default system, which uses no morphological processing.

\item \textbf{lemmatize all}.
        This is the second baseline that uses lemmatized versions
        of the Malay side of the training, development and testing datasets.

\item \textbf{`noisier' channel model}.\footnote{We also tried
the word segmentation model of \newcite{Dyer:lattices:MT:2009}
as implemented in the \emph{cdec} decoder \cite{dyer-EtAl:2010:Demos},
which learns word segmentation lattices from raw text in an unsupervised manner.
Unfortunately, it could not learn meaningful word segmentations for Malay,
and thus we do not compare against it.
We believe this may be due to its focus on word segmentation, which is of limited use for Malay.}
        This is the model of \newcite{Dyer:noisier:channel:2007}.
        It uses 0-1 weights in the lattice and only allows lemmata as alternative wordforms;
        it uses no sentence-level or phrase-level paraphrases.

\end{itemize}

\begin{table}[htb]
\begin{center}
\begin{scriptsize}
\begin{tabular}{llcccc}
\textbf{sent.} & \textbf{system} & \textbf{1-gram} & \textbf{2-gram} & \textbf{3-gram} & \textbf{4-gram}\\
\hline
\hline
  1k & baseline    & 59.78 & 29.60 & 17.36 & 10.46\\
     & paraphrases & 62.23 & 31.19 & 18.53 & 11.35\\
\hline
  2k & baseline    & 64.20 & 33.46 & 20.41 & 12.92\\
     & paraphrases & 66.38 & 35.42 & 21.97 & 14.06\\
\hline
  5k & baseline    & 68.12 & 38.12 & 24.20 & 15.72\\
     & paraphrases & 70.41 & 40.13 & 25.71 & 17.02\\
\hline
 10k & baseline    & 70.13 & 40.67 & 26.15 & 17.27\\
     & paraphrases & 72.04 & 42.28 & 27.55 & 18.36\\
\hline
 20k & baseline    & 73.19 & 44.12 & 29.14 & 19.50\\
     & paraphrases & 73.28 & 44.43 & 29.77 & 20.31\\
\hline
 40k & baseline    & 74.66 & 45.97 & 30.70 & 20.83\\
     & paraphrases & 75.47 & 46.54 & 31.09 & 21.17\\
\hline
 80k & baseline    & 75.72 & 48.08 & 32.80 & 22.59\\
     & paraphrases & 76.03 & 48.47 & 33.20 & 23.00\\
\hline
160k & baseline    & 76.55 & 49.21 & 34.09 & 23.78\\
     & paraphrases & 77.14 & 49.89 & 34.57 & 24.06\\
\hline
320k & baseline    & 77.72 & 50.54 & 35.19 & 24.78\\
     & paraphrases & 78.03 & 51.24 & 35.99 & 25.42\\
\hline
\hline
\end{tabular}
\end{scriptsize}
\end{center}
\caption{
\label{table:bleu:detailed} \textbf{Detailed BLEU $n$-gram precision scores:}
in \%, for different numbers of training sentence pairs,
for \emph{baseline} and \emph{lattice + sent-par + word-par + phrase-par}.}
\end{table}

Our full morphological paraphrasing system is \textbf{lattice + sent-par + word-par + phrase-par}.
We also experimented with some of its
components turned off. \textbf{lattice + sent-par + word-par}
excludes the additional feature from phrase-level paraphrasing.
\textbf{lattice + sent-par} has all the morphologically simpler
derived forms in the lattice during decoding, but their weights are
uniformly set to 0 rather than obtained using pivoting from word
alignments. Finally, in order to compare closely to the `noisier'
channel model, we further limited the morphological variants of
\textbf{lattice + sent-par} in the lattice to lemmata only in
\textbf{lattice + sent-par (orig+lemma)}.

\begin{table}[htb]
\begin{center}
\begin{scriptsize}
\begin{tabular}{@{}l@{ }@{ }lccc@{ }@{ }@{ }c@{ }c@{}}
\textbf{Sent.} & \textbf{System} & \textbf{BLEU} & \textbf{NIST} & \textbf{TER} & \textbf{METEOR} & \textbf{TESLA}\\
\hline
\hline
  1k & baseline    & 23.81 & 6.7013 & 64.50 & 49.26 & 1.6794\\
     & paraphrases & 25.27 & 6.9974 & 63.03 & 52.32 & 1.7579\\
\hline
  2k & baseline    & 27.43 & 7.3790 & 61.03 & 54.29 & 1.8718\\
     & paraphrases & 29.19 & 7.7306 & 59.37 & 57.32 & 2.0031\\
\hline
  5k & baseline    & 31.53 & 8.0992 & 57.12 & 59.09 & 2.1172\\
     & paraphrases & 33.35 & 8.4127 & 55.41 & 61.67 & 2.2240\\
\hline
 10k & baseline    & 33.69 & 8.5314 & 55.24 & 62.26 & 2.2656\\
     & paraphrases & 35.23 & 8.7564 & 53.60 & 63.97 & 2.3634\\
\hline
 20k & baseline    & 36.68 & 8.9604 & 52.56 & 64.67 & 2.3961\\
     & paraphrases & 37.46 & 9.0941 & 52.16 & 66.42 & 2.4621\\
\hline
 40k & baseline    & 38.49 & 9.3016 & 51.20 & 66.68 & 2.5166\\
     & paraphrases & 39.00 & 9.4184 & 50.68 & 67.60 & 2.5604\\
\hline
 80k & baseline    & 40.53 & 9.6047 & 49.88 & 68.77 & 2.6331\\
     & paraphrases & 40.95 & 9.6289 & 49.09 & 69.10 & 2.6628\\
\hline
160k & baseline    & 41.80 & 9.7479 & 48.97 & 69.59 & 2.6887\\
     & paraphrases & 42.30 & 9.8062 & 48.29 & 69.62 & 2.7049\\
\hline
320k & baseline    & 43.02 & 9.8974 & 47.44 & 70.23 & 2.7398\\
     & paraphrases & 43.73 & 9.9945 & 47.07 & 70.87 & 2.7856\\
\hline
\hline
\end{tabular}
\end{scriptsize}
\end{center}
\caption{
\label{table:scores} \textbf{Results for different evaluation measures:}
for \emph{baseline} and \emph{lattice + sent-par + word-par + phrase-par}
(in \% for all measures except for NIST).}
\end{table}

\section{Results and Discussion}
\label{sec:discuss}

The experimental results are shown in Table \ref{table:eval:berita}.

First, we can see that \emph{lemmatize all} has a consistently
disastrous effect on BLEU, which shows that
Malay morphology does indeed contain information
that is important when translating to English.

Second, \newcite{Dyer:noisier:channel:2007}'s \emph{`noisier' channel model}
helps for small datasets only.
It performs worse than \emph{lattice + sent-par (orig+lemma)},
from which it differs in the phrase table only;
this confirms the importance of our sentence-level paraphrasing.

Moving down to \emph{lattice + sent-par},
we can see that using multiple morphological wordforms instead of just lemmata
has a consistently positive impact on BLEU for datasets of all sizes.

Adding weights obtained using word-level pivoting
in \emph{lattice + sent-par + word-par} helps a bit more,
and also using phrase-level paraphrasing weights
yields even bigger further improvements
for \emph{lattice + sent-par + word-par + phrase-par}.

Overall, our morphological paraphrases yield statistically significant improvements ($p < 0.01$)
in BLEU, according to \newcite{Collins:al:2005:sign:test}'s sign test,
for bi-texts as large as 320,000 sentence pairs.
% i.e., for 9.5 million English word tokens.

\textbf{A closer look at BLEU.}
Table \ref{table:bleu:detailed} shows detailed $n$-gram BLEU precision scores for $n$=1,2,3,4.
Our system outperforms the baseline on all precision scores and
for all numbers of training sentences.

\textbf{Other evaluation measures.}
Table \ref{table:scores} reports the results for five evaluation measures:
BLEU and NIST 11b,
TER 0.7.25 \cite{Snover06astudy},
METEOR 1.0 \cite{Lavie:2009:MMA:1743627.1743643},
and TESLA \cite{liu-dahlmeier-ng:2010:WMT}.
Our system consistently outperforms the baseline for all measures.

\begin{table*}[tbh]
\centering
\scriptsize
\begin{tabular}{@{}p{14.4cm}@{}}
  \hline
\texttt{src :} Mercy Relief telah menghantar 17 khemah khas bernilai \$5,000 setiap satu yang boleh menampung kelas seramai 30 pelajar, selain \textbf{bekalan-bekalan lain seperti 500 khemah biasa}, barang makanan dan ubat-ubatan untuk mangsa gempa Sichuan.\\
\texttt{ref1:} Mercy Relief has sent 17 special tents valued at \$5,000 each, that can accommodate a class of 30 students, including \textbf{other aid supplies such as 500 normal tents}, food and medicine for the victims of Sichuan quake.\\
\texttt{base:} mercy relief has sent 17 special tents worth \$5,000 each class could accommodate a total of 30 students, besides \textbf{\emph{other bekalan-bekalan 500 tents as usual}}, foodstuff and medicines for sichuan quake relief.\\
\texttt{para:} mercy relief has sent 17 special tents worth \$5,000 each class could accommodate a total of 30 students, besides \textbf{other supply such as 500 tents}, food and medicines for sichuan quake relief.\\
  \hline
\texttt{src :} Walaupun hidup susah, kami tetap berusaha untuk \textbf{menjalani kehidupan} seperti biasa.\\
\texttt{ref1:} Even though life is difficult, we are still trying to \textbf{go through life} as usual.\\
\texttt{base:} despite the hard life, we will always strive to \textbf{\emph{undergo training}} as usual.\\
\texttt{para:} despite the hard life, we will always strive to \textbf{live} normal.\\
  \hline
\end{tabular}
\caption{{\bf Example translations}.
For each example, we show a source sentence (\texttt{src}),
one of the three reference translations (\texttt{ref1}),
and the outputs of \emph{baseline} (\texttt{base}) and of \emph{lattice + sent-par + word-par + phrase-par} (\texttt{para}).
}
\label{table:examples}
\end{table*}

\textbf{Example translations.}
Table \ref{table:examples} shows two translation examples.
In the first example,
the reduplication \emph{bekalan-bekalan} (`supplies') is an unknown word,
and was left untranslated by the baseline system.
It was not a problem for our system though,
which first paraphrased it as \emph{bekalan} and then translated it as \emph{supply}.
Even though this is still wrong (we need the plural \emph{supplies}),
it is arguably preferable to passing the word untranslated;
it also allowed for a better translation of the surrounding context.

In the second example,
the baseline system translated \emph{menjalani kehidupan} (lit. `go through life') as \emph{undergo training},
because of a bad phrase pair, which 
%was extracted only once and
was extracted from wrong word alignments.
Note that the words \emph{menjalani} (`go through') and \emph{kehidupan} (`life/existence')
%are not rare
%appear in the training bi-text 703 and 735 times, respectively.
%However, they
are %morphologically complex,
%being
derivational forms of \emph{jalan} (`go') and \emph{hidup} (`life/living'), respectively.
Thus, in the paraphrasing system, they were involved in sentence-level paraphrasing,
where the alignments were improved.
While the wrong phrase pair was still available,
the system chose a better one from the paraphrased training bi-text.

%%%%%%%%%%%%%%%%%%%%%%%%%%%%%%%%%%%%%%%%%%%%%%%%%%%
\section{Related Work}
\label{sect:relwork}

Most research in SMT for a morphologically rich \textbf{source} language
has focused on inflected \emph{forms of the same} word.
The assumption is that they would have similar semantics and thus
%the semantics of these forms would be similar to the semantics of the original word,
could have the same translation.
Researchers have used
\emph{stemming} \cite{yang06backoff},
\emph{lemmatization} \cite{Al-onaizan99statisticalmachine,goldwater:mcclosky:2005,Dyer:noisier:channel:2007},
or \emph{direct clustering} \cite{Talbot:modellinglexical:2006}
to identify such groups of words and use them as \emph{equivalence classes}
or as possible \emph{alternatives} in translation.
Frameworks for the simultaneous use of different word-level representations
have been proposed as well
%such as the \emph{factored translation model}
\cite{koehn-hoang:2007:EMNLP-CoNLL2007}.

A second important line of research has focused on \emph{word segmentation},
which is useful for languages like German,
which are rich in \emph{compound words} that are spelled concatenated \cite{Koehn:2003:EMC,yang06backoff},
or like Arabic, Turkish, Finnish, and, to a lesser extent, Spanish and Italian,
where \emph{clitics} often attach to the preceding word \cite{habash-sadat:2006:HLT-NAACL06-Short}.
For languages with more or less regular inflectional morphology like Arabic or Turkish,
another good idea is to segment words into \emph{morpheme sequences},
e.g., prefix(es)-stem-suffix(es),
which can be used instead of the original words \cite{lee:2004:HLTNAACL}
or in addition to them.
This can be achieved using a lattice input
to the translation system \cite{dyer-muresan-resnik:2008:ACLMain,Dyer:lattices:MT:2009}.

Unfortunately, none of these general lines of research suits Malay well,
whose compounds are rarely concatenated,
clitics are not so frequent,
and morphology is mostly derivational,
and thus likely to generate words whose semantics substantially differs from the semantics of the original word.
Therefore, we cannot expect the existence of equivalence classes:
it is only occasionally that two derivationally related wordforms
would share the same target language translation.
Thus, instead of looking for
equivalence classes,
we have focused on the pairwise relationship between derivationally related wordforms,
which we treat as \emph{potential paraphrases}.

Our approach is an extension of the \emph{`noisier' channel model} of \newcite{Dyer:noisier:channel:2007}.
He starts by generating separate word alignments
for the original training bi-text
and for a version of it where the source side has been lemmatized.
Then, the two bi-texts and their word alignments are concatenated
and used to build a phrase table.
Finally, the source sides of the development and the test datasets
are converted into confusion networks where
additional arcs are added for word lemmata.
The arc weights are set to 1 for the original wordforms
and to 0 for the lemmata.
In contrast, we provide \emph{multiple} paraphrasing alternatives for each morphologically complex word,
including derivational forms that occupy intermediary positions between the original wordform and its lemma.
Note that some of those paraphrasing alternatives are \emph{multi-word},
and thus we use a \emph{lattice} instead of a confusion network.
Moreover, we give \emph{different weights} to the different alternatives rather then assigning them all 0.

Second, our work is related to that of \newcite{dyer-muresan-resnik:2008:ACLMain},
who use a lattice to add a single alternative clitic-segmented version of the original word for Arabic.
However, we provide \emph{multiple} alternatives.
We also include \emph{derivational forms} in addition to clitic-segmented ones,
and we give \emph{different weights} to the different alternatives (instead of 0).

Third, our work is also related to that of \newcite{Dyer:lattices:MT:2009},
who uses a lattice to add multiple alternative segmented versions of the original word
for German, Hungarian, and Turkish.
However, we focus on \emph{derivational morphology} rather than on clitics and inflections,
add \emph{derivational forms} in addition to clitic-segmented ones,
and use \emph{cross-lingual word pivoting} to estimate paraphrase probabilities.

Finally, our work is related to that of \newcite{callison06paraphrase},
who use cross-lingual pivoting to generate phrase-level paraphrases with corresponding probabilities.
However, our paraphrases are derived through \emph{morphological analysis};
thus, we do not need corpora in additional languages.

\section{Conclusion and Future Work}
\label{sec:conclusion}

We have presented a novel approach
to translating from a morphologically complex language,
which uses paraphrases and paraphrasing techniques
at three different levels of translation:
word-level, phrase-level, and sentence-level.
Our experiments translating from Malay,
whose morphology is mostly derivational,
into English have shown significant improvements over rivaling approaches
based on several automatic evaluation measures.

In future work,
we want to improve the probability estimations for our paraphrasing models.
We also want to experiment with other morphologically complex languages
and other SMT models.

\section*{Acknowledgments}

This work was supported by research grant POD0713875. We would like to
thank the anonymous reviewers for their detailed and constructive
comments, which have helped us improve the paper.

\bibliographystyle{acl}
\bibliography{acl-hlt2011}

\end{document}